\documentclass[12pt,a4paper]{article}
\usepackage[margin=1in]{geometry}
\usepackage{amsmath,amssymb,amsfonts}
\usepackage{graphicx}
\usepackage{xcolor}
\usepackage[colorlinks=true,allcolors=blue!50!black]{hyperref}
\usepackage{natbib}
\usepackage{booktabs}

\title{Verifying Human-Written Reasoning References}

\author{
Ali Ansari \qquad Farhana Ferdousi Liza\\[0.4em]
School of Computing Sciences, University of East Anglia\\
}
\date{}

\begin{document}

\maketitle

\begin{abstract}
Evaluating the reasoning of AI systems often relies on human-written reasoning chains as references, yet the logical support for their individual steps is rarely verified. Although prior neuro-symbolic approaches have formalized and checked reasoning, their reported evaluations do not systematically assess whether step conclusions in human-written reference derivations are entailed by their cited premise sets or investigate the sources of failures in those references. To address these limitations, we present a novel reference-centred assessment method for human-written derivations that combines entailment checking against cited premise sets, full-premise rechecking, and failure attribution to examine the reliability of these derivations and their formal representations. The proposed method is designed for reasoning datasets containing human-written derivations with explicitly cited premise sets. We evaluate it on P-FOLIO, using a large language model (LLM) to translate step conclusions into first-order logic (FOL) and the Z3 SMT solver to check entailment. Each translated step is first checked for entailment from the premises or earlier derivation steps explicitly referenced by the annotator. Steps that fail this check are then rechecked against the complete set of original premises to distinguish insufficient recorded dependencies from claims that are not entailed even by the full premise set.

Of 386 derivations examined, 202 (52.3\%) meet all our checking criteria, meaning that every translated step is entailed by its stated support and the required target is established under the available FOL representations. Among 230 steps not entailed by their stated support, 191 become entailed when checked against the complete original premises, indicating that most initial failures reflect incomplete or mismatched support rather than unsupported claims. Manual analysis of the remaining 39 cases identifies 21 human reasoning errors, 11 FOL translation errors, and 7 dataset annotation errors. A separate manual audit of 50 randomly sampled steps that passed the entailment checks found 46 faithful FOL translations (92\%). These findings suggest that disagreement with a human-written reasoning reference may originate in the reference or its formal representation rather than in the AI system being evaluated. By identifying potentially unsupported steps and distinguishing sources of failure, our approach can help researchers prioritize reference review and interpret disagreements between models and human-written references more cautiously.
\end{abstract}

\section{Introduction}

The reliability of Artificial Intelligence (AI) evaluation depends not only on the systems being assessed but also on the quality of the reference data used to evaluate them. This is particularly important in reasoning tasks, where researchers increasingly examine the intermediate reasoning processes leading to an answer rather than focusing only on the final result~\citep{prasad-etal-2023-receval,han2024p}. Human-written reasoning chains can provide useful references for such evaluations, but their reliability depends on whether the claims and justifications within those chains accurately represent the underlying logical relationships. When the reference reasoning is incomplete or incorrect, disagreement with a model-generated reasoning chain becomes difficult to interpret.

Previous research has shown that human-annotated explanations may not constitute logically valid arguments, even when they appear coherent~\citep{valentino-etal-2021-natural}. Errors have also been identified in FOLIO's released first-order logic annotations~\citep{brunello2026fixing}.  These findings raise concerns about treating human-written reasoning and its formal representations as inherently reliable. 
Consequently, disagreement with a model-generated explanation may reflect problems in the reference itself rather than necessarily indicating incorrect reasoning by the model.

The problem extends beyond the correctness of final answers. A human-written derivation may reach the expected conclusion while leaving gaps in the justification of step conclusions. Similarly, errors in a formal annotation may affect how the original reasoning is interpreted. These possibilities make the reliability of human-written reasoning references an important consideration for research based on annotated datasets. We define \emph{epistemic risk} as the extent to which errors in human annotations become entrenched as unquestioned ground truth in a dataset, potentially undermining the reliability of knowledge claims derived from that dataset.

Logical reasoning benchmarks provide a particularly relevant setting for examining this concern. FOLIO contains natural-language reasoning problems accompanied by first-order logic annotations~\citep{han2024folio}. P-FOLIO extends this dataset with human-written derivations that record intermediate reasoning steps and their dependencies on original premises or earlier steps~\citep{han2024p}. These additional explanations provide more information than final-answer labels alone, but they also introduce questions about whether the recorded reasoning adequately justifies the conclusions being drawn. The presence of formal annotations does not, by itself, establish that every human-written derivation step is logically supported.

Existing research has examined several aspects of logical reasoning reliability. \citet{ganguly2024pot} introduce Proof of Thought, which converts structured LLM outputs into first-order logic for automated verification, while \citet{mensfelt2025lelma} combine logical validation with feedback to improve LLM-generated reasoning. \citet{pei2025fover} use LLM-assisted FOL translation and Z3 to verify reasoning texts, including on FOLIO, and identify annotation errors. Similarly, \citet{feng2026vericot} formalize and verify individual steps in LLM-generated reasoning chains and examine their supporting premises. \citet{huang2026forex} use Lean4 to assess whether formalized LLM-generated explanations follow from encoded premises and compare verification outcomes with human-annotated labels. In a complementary direction, \citet{brunello2026fixing} examine errors in FOLIO's released FOL annotations and labels. Although these studies address formal reasoning verification and annotation reliability, they do not report a systematic assessment of whether human-written derivations in P-FOLIO follow from their explicitly recorded dependencies, together with an analysis of failure sources and translation faithfulness.

These considerations motivate the central research question of this study: to what extent are the conclusions of human-written derivation steps entailed by their explicitly cited premise sets, and what kinds of failures or mismatches arise from the reasoning, its cited premise set, and its formal representations?  Our work focuses on the reliability of human-written reference derivations and distinguishes problems arising from the reference reasoning itself, its recorded dependencies, and the formal representations used to assess it. By examining these sources of uncertainty, we aim to provide a clearer understanding of when reasoning references can support meaningful evaluation and when disagreements with them should be interpreted more cautiously.

\section{Background}

Benchmark-based evaluation provides a common way to compare machine learning systems, but the results depend on the tasks, datasets, and evaluation metrics used. \citet{dehghani2021benchmark} discuss how benchmark selection can influence comparisons between machine learning methods, while \citet{ribeiro-etal-2020-beyond} show that standard test accuracy may fail to reveal important behavioural weaknesses in NLP models. These findings highlight the importance of understanding what an evaluation actually measures. In reasoning tasks, final-answer accuracy indicates whether a model reaches the expected conclusion, but does not necessarily establish whether the intermediate reasoning leading to that conclusion is logically valid. Evaluating reasoning therefore requires considering not only the final answer but also the information used to justify it.

The quality of reference data is another important factor in evaluation. Dataset annotations may include labels, supporting evidence, or formal representations that determine how a system's output is judged. \citet{gebru2021datasheets} emphasize the importance of documenting dataset construction and annotation decisions, while \citet{sambasivan2021datacascades} examine how problems originating in data preparation can affect downstream machine learning systems. In logical reasoning datasets, annotation quality is particularly important because the meaning of a statement depends on its logical structure. An omitted condition, an incorrectly represented relation, or a change in quantifier scope can alter what follows from the available information. This makes it important to distinguish the correctness of a reference annotation from the correctness of the system being evaluated against it.

Reasoning benchmarks increasingly include human-written explanations and rationales alongside final-answer labels. For example, ERASER provides human-annotated evidence for evaluating model explanations~\citep{deyoung-etal-2020-eraser}. However, providing an explanation is not the same as establishing that its conclusion logically follows from the available evidence. \citet{carton-etal-2020-evaluating} investigate the quality of human rationales and highlight limitations in treating them as reliable explanations. More directly, \citet{valentino-etal-2021-natural} examine the logical validity of human-annotated explanations and find that explanations which appear plausible may nevertheless contain invalid arguments. These studies illustrate an important distinction between an explanation that appears relevant or convincing and a derivation whose individual claims are logically supported. For step-by-step reasoning, this distinction also concerns the relationship between each step conclusion and the information used to justify it.

Formal logic provides a way to represent and examine these relationships explicitly. First-order logic (FOL) expresses statements using predicates, logical connectives, and quantifiers, allowing relationships between premises and conclusions to be represented formally. Given a set of premises $S$ and a candidate conclusion $C$, logical entailment is written as
\[
S \models C,
\]

 
meaning that every interpretation satisfying $S$ also satisfies $C$. Entailment can be checked by testing whether the set
\[
S \cup \{\neg C\}
\]
is unsatisfiable. Satisfiability modulo theories (SMT) solvers, such as Z3, support automated reasoning over formal constraints and are widely used in verification~\citep{ge2009quantified,barrett2009smt,de2008z3}. In classical logic, a set of premises is logically inconsistent if no interpretation satisfies all of them simultaneously. Such a set entails every formula, a property known as the principle of explosion (\emph{ex falso quodlibet})~\citep{shapiro2026classical}. Consequently, an unsatisfiable entailment query may result from inconsistent support rather than a meaningful justification for the claim. Our procedure therefore checks the satisfiability of the stated support separately and does not count inconsistent support as a successful entailment check. Furthermore, formal entailment establishes a relationship between the supplied formulas; it does not by itself guarantee that those formulas faithfully represent their original natural-language statements.

FOLIO and P-FOLIO provide a setting in which natural-language reasoning can be examined alongside formal representations. We focus on P-FOLIO because it combines human-written intermediate derivation steps, explicitly recorded dependencies, and FOL annotations for the original problems. The alternative benchmarks considered in Section~\ref{sec:dataset} do not provide this combination of features required for our analysis. FOLIO contains natural-language premises and conclusions paired with first-order logic annotations and labels indicating whether a conclusion is entailed, contradicted, or undetermined by the premises~\citep{han2024folio}. P-FOLIO extends this dataset with human-written, step-by-step derivations~\citep{han2024p}. Each derivation step contains a natural-language claim, references to the original premises or earlier derivation steps on which it depends, and an inference-rule label. Unlike the original premises and conclusions, these intermediate derivation steps are not accompanied by corresponding FOL formulas. This distinction is important because the dataset provides both the logical representation of the original problem and the human-recorded dependencies within its reasoning chain. Together, these components provide a basis for studying the relationship between formal annotations, step conclusions, and their cited premise sets.

\section{Method}\label{sec:method}

This section defines the formal entailment criteria used to evaluate individual human-written derivation steps and complete reasoning chains.

\subsection{Analytical Method}

We operationalize reference reliability in terms of logical alignment across four connected representations: the natural-language problem, its first-order logic annotations, the human-written derivation steps, and the premises or earlier derivations cited as support for each step. These representations jointly determine how the reasoning encoded in a problem instance is interpreted. A mismatch at any layer can therefore change the outcome of subsequent evaluation. For example, a failed step-level entailment check may arise from an invalid human inference, omitted relevant premises or earlier derivation steps in the cited premise set, or a mismatch between the natural-language statement and its FOL representation.

At the level of an individual derivation step, we assess entailment over FOL representations. For each derivation step $i$, let $c_i$ denote the FOL formula representing its step conclusion, and let $S_i$ denote the set of FOL formulas corresponding to the original premises or earlier derivation steps explicitly cited by the annotator. The step passes the entailment check when
\[
S_i \models c_i.
\]
The step conclusion represented by $c_i$ may coincide with the problem-level target, but the two are distinguished in our notation. The test uses only the cited premise set $S_i$, without adding uncited original premises.

Step-level entailment alone is not sufficient to assess a complete reasoning chain. We use $T$ to denote the problem-level target: for an instance labelled True, $T$ is the released FOL formula for the stated conclusion; for an instance labelled False, $T$ is its negation. We treat a derivation as meeting our checking criteria only when every translated step is entailed by its cited premise set and the required target $T$ is established under the available FOL representations. This distinguishes establishing the problem-level target from justifying each step conclusion: a derivation may establish $T$ while still containing steps not entailed by their cited support.

A derivation step whose translated step conclusion is not entailed by its cited premise set does not necessarily contain an invalid human inference. We therefore recheck the corresponding FOL formula against the complete original premise set. Cases that remain non-entailed are manually examined to distinguish human reasoning errors, FOL translation errors, and dataset annotation errors.

\subsection{Dataset}
\label{sec:dataset}

The checking procedure requires a dataset that exposes the reasoning reference at several connected levels. In particular, the problem instances must provide natural-language premises and conclusions together with corresponding formal representations, so that their logical content can be examined mechanically. The dataset must also contain human-written reasoning chains rather than only final labels, since our objective is to assess whether step conclusions are entailed by their cited premise sets. Finally, the derivation steps must identify the premises or earlier steps on which they depend. This last property is particularly important because our step-level analysis initially evaluates each claim against the support explicitly identified by the human annotator, rather than only against the complete problem context.

Existing logical reasoning benchmarks expose different subsets of these requirements. ProofWriter~\citep{tafjord2021proofwriter} and RuleTaker~\citep{clark2020transformers} provide explicit reasoning structures, but their problems and reasoning are generated from controlled templates rather than supplied as natural-language derivations written by human annotators. ProntoQA~\citep{saparov2023language} also provides inspectable reasoning chains, but is based on synthetic problems constructed from a restricted set of inference patterns. LogicNLI~\citep{tian2021logicnli} is grounded in first-order logic but does not provide human-written intermediate derivations with explicit step dependencies for the type of verification performed in this study. LogiQA~\citep{liu2020logiqa} and ReClor~\citep{yu2020reclor} contain human-authored logical reasoning problems, but do not provide the combination of formal annotations and explicit human-written derivation steps required by our method. These differences make P-FOLIO particularly suitable for examining the reliability of reasoning references rather than only their final answers.

We therefore use P-FOLIO~\citep{han2024p}, which extends FOLIO~\citep{han2024folio}. FOLIO provides natural-language premises and target conclusions together with corresponding first-order logic (FOL) annotations for deductive reasoning problems. P-FOLIO augments these problem instances with human-written natural-language derivations. Each derivation consists of an ordered sequence of steps in which the annotator states a step conclusion, identifies its cited premise set, and provides an inference-rule label. The derivation steps themselves are provided only in natural language and are not accompanied by corresponding FOL formulas. This creates a useful setting for our analysis: the original problem already has a formal representation, while the intermediate human reasoning must first be formalized before its logical validity can be checked automatically. Although the original P-FOLIO study included human cross-checking of the annotated proofs and logic-solver checks for some complex inference rules~\citep{han2024p}, it did not report a systematic FOL entailment analysis of each released derivation step against its recorded dependencies.

This combination allows several layers of the reference to be examined within the same problem instance. The natural-language premises and conclusion can be compared with their released FOL annotations; each human-written derivation step can be checked against the premises or earlier derivations it cites; and the complete derivation can be assessed against the target that the problem label requires it to establish. Failures of the entailment checks are therefore not automatically attributed to human reasoning, particularly given previously reported errors in FOLIO's released FOL annotations~\citep{brunello2026fixing}.

We restrict derivation-level checking to instances labelled \emph{True} or \emph{False}, for which the required target is the stated conclusion or its negation, respectively. Instances labelled \emph{Uncertain} are excluded because our derivation-level checking criterion requires a definite target.

\subsection{Pilot Study on P-FOLIO}\label{sec:exp_design}

The proposed entailment-checking method is not specific to P-FOLIO. It can be applied to datasets containing human-written derivations with explicitly cited premise sets for their step conclusions and a defined problem-level target. In this study, we apply the method to P-FOLIO, using its released FOL annotations and translating the step conclusions into FOL. This subsection describes the translation and entailment-checking pipeline, while the experimental results are presented in Section~\ref{sec:results}. Applying the method to datasets with different formal representations may require adapting the translation and parsing stages.

\subsubsection{Checking Human-Written Derivations}
\label{sec:pipeline}

Each P-FOLIO derivation step states a step conclusion in natural language and cites original premises or earlier derivation steps~\citep{han2024p}. We translate each step conclusion into FOL and check whether the resulting formula is entailed by its cited premise set, following the criteria defined in Section~\ref{sec:method}.

The following P-FOLIO problem illustrates the structure used by the pipeline.

\paragraph{Example illustration: P-FOLIO Story 59, conclusion index 2}

\begin{quote}
\textbf{Premises.} \emph{Natural language, with the corresponding FOL annotations.}

\begin{enumerate}\itemsep2pt
\item Some monitors equipped in the lab are produced by the company named AOC.
\[
\exists x\,\exists y\,
(\mathrm{LabMonitor}(x)
\land
\mathrm{AOC}(x)
\land
\neg(x=y)
\land
\mathrm{LabMonitor}(y)
\land
\mathrm{AOC}(y))
\]

\item All monitors equipped in the lab are cheaper than their original prices.
\[
\forall x\,
(\mathrm{LabMonitor}(x)
\rightarrow
\mathrm{Discounted}(x))
\]

\item If a monitor is cheaper than its original price, then its resolution is 1080p.
\[
\forall x\,
(\mathrm{Discounted}(x)
\rightarrow
\mathrm{A1080p}(x))
\]

\item If a monitor has a resolution of 1080p, then it does not support the type-c port.
\[
\forall x\,
(\mathrm{A1080p}(x)
\rightarrow
\neg\mathrm{TypeC}(x))
\]

\item LG34 is equipped in the lab.
\[
\mathrm{LabMonitor}(\mathrm{lg\mbox{-}34})
\]
\end{enumerate}

\vspace{6pt}

\textbf{Conclusion.} LG34 is not with a resolution of 1080p.
\emph{(label: False)}
\[
\neg\mathrm{A1080p}(\mathrm{lg\mbox{-}34})
\]

\vspace{8pt}

\textbf{Derivation.}
\emph{P-FOLIO natural-language steps with their stated support and inference-rule labels.}

\noindent
\parbox[t]{0.72\linewidth}{
\texttt{D1: From (5,2): LG34 machine is cheaper than its original price.}}
\hfill
\parbox[t]{0.24\linewidth}{
\raggedleft\emph{[universal instantiation]}}
\\[6pt]

\parbox[t]{0.72\linewidth}{
\texttt{D2: From (D1,3): LG34 has a resolution of 1080p.}}
\hfill
\parbox[t]{0.24\linewidth}{
\raggedleft\emph{[universal instantiation]}}
\end{quote}

In this example, D1 cites premises~5 and~2, while D2 cites the earlier derivation step D1 together with premise~3. The example therefore contains both types of dependency used in our verification procedure: direct dependence on original premises and dependence on previously established step conclusions. Because neither D1 nor D2 is supplied with a FOL representation in P-FOLIO, their natural-language claims must first be formalized before logical entailment can be checked automatically.

\begin{figure}[!htb]
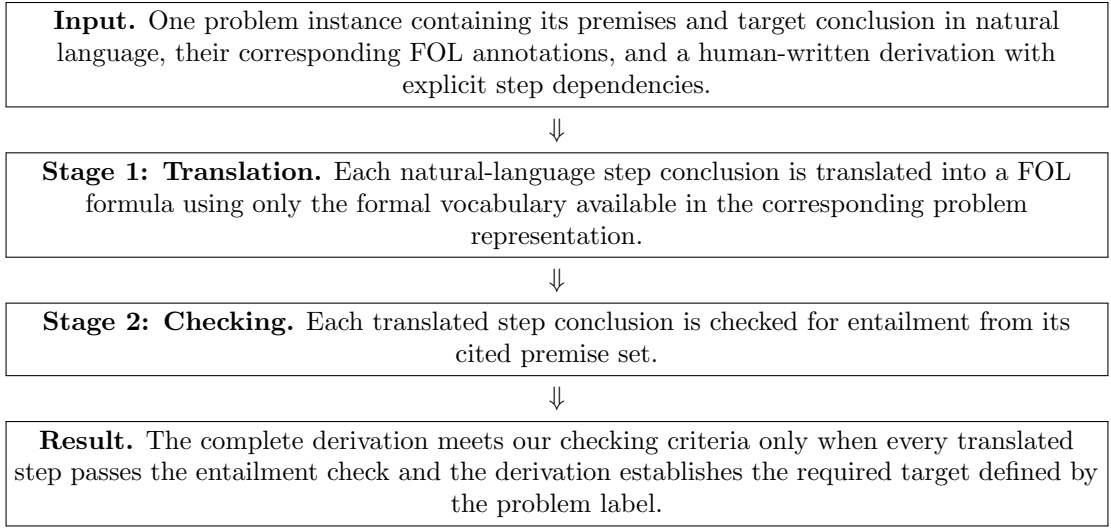

\centering
\footnotesize

\fbox{\parbox{0.9\textwidth}{\centering
\textbf{Input.} One problem instance containing its premises and target conclusion in natural language, their corresponding FOL annotations, and a human-written derivation with explicit step dependencies.}}
\\[3pt]

$\Downarrow$
\\[3pt]

\fbox{\parbox{0.9\textwidth}{\centering
\textbf{Stage 1: Translation.} Each natural-language step conclusion is translated into a FOL formula using only the formal vocabulary available in the corresponding problem representation.}}
\\[3pt]

$\Downarrow$
\\[3pt]

\fbox{\parbox{0.9\textwidth}{\centering
\textbf{Stage 2: Checking.} Each translated step conclusion is checked for entailment from its cited premise set.}}
\\[3pt]

$\Downarrow$
\\[3pt]

\fbox{\parbox{0.9\textwidth}{\centering
\textbf{Result.} The complete derivation meets our checking criteria only when every translated step passes the entailment check and the derivation establishes the required target defined by the problem label.}}

\caption{Pipeline for assessing step-level entailment in human-written reasoning derivations. The present implementation instantiates the Translation and Checking stages using P-FOLIO, o3-mini, and Z3.}
\label{fig:reference-pipeline}
\end{figure}

The pipeline comprises two stages: \emph{Translation}, which represents each natural-language step conclusion as a FOL formula, and \emph{Checking}, which tests whether that formula is entailed by its cited premise set. Derivation steps are read directly from the structured P-FOLIO data, so no separate extraction stage is required. Figure~\ref{fig:reference-pipeline} summarizes the procedure.

\paragraph{Translation.}

Since P-FOLIO does not provide FOL representations for its derivation steps, each natural-language step conclusion must be formalized before entailment checking. We use the LLM o3-mini for this translation in the present implementation. The translation prompt instructs the model to use only predicates and constants already present in the problem's FOL annotations. This is intended to keep the translated claims grounded in the same formal vocabulary as the premises and target conclusion. However, the model may not always follow this instruction, and vocabulary violations are considered in our subsequent error analysis.

We selected o3-mini because it obtained the highest natural-language-to-FOL accuracy among the models evaluated by \citet{brunello2026do}. This prior result motivates its use as the formalizer in our implementation, but does not establish that every translation produced for P-FOLIO derivation steps is correct. Translation therefore remains a possible source of verification failure and is examined explicitly in our later analysis. When a step cannot be represented faithfully using the available vocabulary, no approximate or repaired translation is introduced; the step is instead recorded as having no usable translation.

For Story~59, D1 is translated as
\[
\mathrm{Discounted}(\mathrm{lg\mbox{-}34}),
\]
and D2 as
\[
\mathrm{A1080p}(\mathrm{lg\mbox{-}34}).
\]
Both formulas use only predicates and constants already available in the original problem representation.

\paragraph{Faithful translation.}

The formalization stage is instructed to preserve the content of the human-written derivation step rather than to repair it. This distinction is necessary because automatically repairing the logical content of a human-written claim during translation could cause an invalid reference step to pass the subsequent verification stage. When the content of a step cannot be represented faithfully using the available FOL vocabulary, the pipeline records the absence of a usable translation rather than introducing information that is not present in the original step.

\paragraph{Checking.}

For derivation step $i$, we test whether its cited premise set $S_i$ entails the FOL formula $c_i$ representing its step conclusion. This is checked by determining whether
\[
S_i \cup \{\neg c_i\}
\]
is satisfiable. If this set is unsatisfiable, then
\[
S_i \models c_i.
\]
In the present implementation, this satisfiability test is performed using Z3~\citep{de2008z3}. The use of a satisfiability solver allows the entailment check to be implemented mechanically for each translated derivation step.

The check is deliberately restricted to the support explicitly identified by the human-written derivation. It does not initially use every premise available in the problem. This distinction allows the pipeline to test whether the cited premise set entails each step conclusion, rather than merely whether that conclusion is entailed by the complete original premise set.

Several structural conditions are evaluated before the entailment query is sent to Z3. A step is not checked normally when it refers to a missing earlier derivation step, refers to itself, refers to a later step, or reuses an existing step identifier. Similarly, if a step depends on an earlier derivation step that has failed the checking procedure, the dependent step cannot be established through that step conclusion. Support that is internally inconsistent is also recorded separately, since an inconsistent set would entail every formula trivially and would therefore make a successful entailment result uninformative.

Story~59 illustrates the checking procedure. For D1, the stated support consists of premise~5,
\[
\mathrm{LabMonitor}(\mathrm{lg\mbox{-}34}),
\]
and premise~2,
\[
\forall x\,
(\mathrm{LabMonitor}(x)
\rightarrow
\mathrm{Discounted}(x)).
\]

The corresponding entailment query tests whether
\[
\mathrm{LabMonitor}(\mathrm{lg\mbox{-}34})
\land
\forall x\,
(\mathrm{LabMonitor}(x)
\rightarrow
\mathrm{Discounted}(x))
\land
\neg\mathrm{Discounted}(\mathrm{lg\mbox{-}34})
\]
is satisfiable. It is unsatisfiable, so D1 follows from its stated support.

D2 is checked using the previously entailed FOL claim from D1 together with premise~3. The query is
\[
\forall x\,
(\mathrm{Discounted}(x)
\rightarrow
\mathrm{A1080p}(x))
\land
\mathrm{Discounted}(\mathrm{lg\mbox{-}34})
\land
\neg\mathrm{A1080p}(\mathrm{lg\mbox{-}34}).
\]
This formula is also unsatisfiable, so D2 passes the entailment check. Because Story~59 is labelled False, the required target is the negation of the stated conclusion,
\[
\mathrm{A1080p}(\mathrm{lg\mbox{-}34}),
\]
which is established by D2. The complete derivation therefore meets our checking criteria.

The example also shows why verification is performed against the stated support of each step. D1 requires both premises~5 and~2: premise~5 establishes that LG34 is a lab monitor, while premise~2 establishes that every lab monitor is discounted. If D1 cited only premise~2, the translated claim would not follow from the information explicitly supplied by that step, even though the necessary additional information exists elsewhere in the problem.

\paragraph{Interpreting failed checks.}

A failed entailment check is treated as a verification failure rather than as immediate evidence that the human-written reasoning is logically incorrect. After the initial step-level check, claims that are not entailed by their stated support are checked again against the complete set of original premises. This wider test distinguishes translated claims entailed by the complete original premise set but not by their cited premise sets from those not entailed by the complete original premise set. The latter cases are then examined manually to determine whether the mismatch originates in the human-written reasoning, the automatic FOL translation, or the released dataset annotation.

\paragraph{Scope of entailment checking.}

Our checking criteria assess whether the FOL translation of each derivation step is entailed by its cited premise set and whether the required target is established under the representations used by the pipeline. These are formal entailment checks, but they do not constitute end-to-end formal verification of the original natural-language reasoning, because the LLM-generated translations are not independently guaranteed to preserve the intended meaning. The procedure also does not verify whether the inference-rule label attached to a derivation step is correct: the checking stage evaluates logical entailment rather than the schema of the named inference rule. Failure to meet the criteria does not necessarily indicate incorrect natural-language reasoning. It may arise from incomplete dependencies, structural problems, limitations of the available FOL vocabulary, translation errors, or mismatches in the released formal annotations. These possibilities are examined in the subsequent results and analysis.

\paragraph{Manual translation-faithfulness audit.}

To assess translation reliability beyond the failed entailment checks, we manually reviewed a random sample of 50 derivation steps from the 1,075 steps that Z3 classified as entailed by their stated support. For each sampled step, we compared its natural-language claim with the automatically generated FOL translation, using the original premises and formal vocabulary as context. We classified translations as faithful, weaker, stronger, or changed according to whether they preserved, weakened, strengthened, or otherwise altered the intended meaning. This assessment focused on semantic correspondence between the natural-language claim and its FOL representation, independently of the entailment outcome.

\paragraph{Problem-level annotation checks.}

Before applying the derivation-checking pipeline, we check whether the released FOL representation of each problem can support the evaluation itself. We first test whether the supplied FOL annotations can be parsed and whether the original premises entail the target required by the assigned True or False label. Instances that require manual or automatic repair to obtain the expected entailment are not included in the working set. For annotations that parse but do not support their assigned label, we additionally apply rule-based diagnostics to identify recurring mismatch patterns, including absent predicates or constants, inconsistent symbol naming, unbound variables, and cases in which a single conclusion-side symbol substitution would restore entailment. These diagnostics are used only to characterize the failures; no proposed correction is applied to the data used by the checking pipeline.

Further implementation details and the complete translation prompt are
provided in Appendix~\ref{app:implementation} and
Appendix~\ref{app:translation-prompt}.

\section{Experiments and Results}
\label{sec:results}

We begin with the 871 P-FOLIO problem instances that contain a human-written derivation and apply the dataset-filtering criteria described below, resulting in a working set of 386 instances. We then apply the Translation and Checking stages defined in Section~\ref{sec:pipeline}, followed by derivation-level checking. All results for the main working set use the released FOL annotations and automatically produced FOL translations without manual correction. Instances whose annotations require parenthesis repair are excluded rather than repaired, because such a change may alter quantifier scope and would require an assumption about the intended formal representation.

\subsection{Construction of the Evaluation Set}
\label{sec:results-pool}

Derivation-level checking requires both a definite target and a usable formal representation of the problem. Following FOLIO's label scheme, an instance is labelled True when the premises entail the stated conclusion, False when they entail its negation, and Uncertain when neither the conclusion nor its negation is entailed~\citep{han2024folio}. We therefore exclude Uncertain instances because they provide no definite target for derivation-level checking. We also exclude annotations that cannot be parsed as first-order logic and parseable annotations whose premises do not support the assigned True or False label. Finally, we exclude cases in which the expected entailment is obtained only after adding or removing a parenthesis, since such a change may alter quantifier scope and would require an assumption about the intended annotation.

Table~\ref{tab:funnel} summarizes these filtering stages. Starting from the 871 P-FOLIO instances with a human-written derivation, we retain instances with a True or False label, whose FOL annotations parse, whose released FOL representation supports the assigned label, and that require no parenthesis repair. This results in a working set of 386 instances.

\begin{table}[!htb]
\centering
\small
\begin{tabular}{lr}
\toprule
Stage & Instances \\
\midrule
FOLIO $\times$ P-FOLIO problem instances & 1{,}434 \\
With a human-written derivation & 871 \\
True or False label & 868 \\
FOL annotation parses & 763 \\
FOL annotation supports the assigned label & 409 \\
No parenthesis repair required & \textbf{386} \\
\bottomrule
\end{tabular}
\caption{Construction of the evaluation set. The final 386 instances are checked using the released FOL annotations without annotation repair or manual correction.}
\label{tab:funnel}
\end{table}

\subsection{Annotation-Level Mismatches}
\label{sec:results-annotations}

The 354 parseable annotations whose original premise sets do not entail the targets required by their assigned labels provide a separate view of annotation quality. The rule-based diagnostics described in Section~\ref{sec:pipeline} identify one of the predefined mismatch patterns in 208 cases. These consist of a conclusion predicate absent from the premises (75 cases), inconsistent symbol naming (45), a conclusion constant absent from the premises (37), an unbound variable (28), and 23 cases in which substituting a single conclusion-side symbol is sufficient to restore the expected entailment.

The remaining 146 cases are not explained by these diagnostics. We do not interpret them as correct annotations; rather, their source of disagreement remains unresolved by the diagnostic checks used here. Likewise, the 23 proposed symbol substitutions are treated only as diagnostic observations. None of these changes is applied when constructing the 386-instance working set.

\subsection{Step-Level Entailment Checking}
\label{sec:results-steps}

The 386 derivations contain 1{,}514 human-written steps. Translation produces a usable FOL translation for 1{,}371 of them, corresponding to 90.6\%. Most of the untranslated cases arise because the claim cannot be represented faithfully using the FOL vocabulary available for that problem; only one generated formula fails to parse.

Not every translated claim reaches the entailment test. Sixty-six steps are stopped before Z3 can evaluate the claim against its stated support. Most of these cases, 48 steps, refer to an earlier derivation identifier that is missing from the released derivation. The remaining 18 involve other structural conditions, including self-reference, references to later steps, repeated identifiers, dependence on an earlier failed step, or inconsistent support.

Z3 therefore evaluates 1{,}305 translated steps. Of these, 1{,}075 are entailed by their stated support, while 230 are not. Table~\ref{tab:steps} summarizes the complete step-level outcome. Non-entailment from the cited premise set does not establish non-entailment from the complete original premise set. We therefore examine these 230 cases under a wider support condition in the following subsection.

\begin{table}[!htb]
\centering
\small
\begin{tabular}{lr}
\toprule
Step-level outcome & Steps \\
\midrule
Total derivation steps & 1{,}514 \\
Usable FOL translation & 1{,}371 \\
No usable FOL translation & 143 \\
\midrule
Translated but not checked with Z3 & 66 \\
\quad Missing referenced derivation step & 48 \\
\quad Other structural cases & 18 \\
\midrule
Checked with Z3 & 1{,}305 \\
\quad Entailed by stated support & 1{,}075 \\
\quad Not entailed by stated support & 230 \\
\bottomrule
\end{tabular}
\caption{Step-level translation and entailment-checking outcomes. Among the 1{,}305 steps evaluated by Z3, 82.4\% are entailed by their stated support and 17.6\% are not.}
\label{tab:steps}
\end{table}

\subsection{What Explains the Failed Entailment Checks?}
\label{sec:results-full-context}

To further examine the 230 translated claims not entailed by their cited premise sets, we repeat the entailment test using the complete set of original premises for each problem. Earlier derivation steps are excluded from this analysis so that the result depends only on the original problem representation. All premise sets in the working set are satisfiable, avoiding trivial entailment from inconsistency.

This wider check changes the interpretation of most failures. Of the 230 claims that are not entailed by their stated support, 191 (83.0\%) are entailed by the complete original premise set, while the remaining 39 (17.0\%) are not entailed by those premises. A failed step-level check therefore often reflects an incomplete cited premise set rather than non-entailment from the complete original premise set. This may take the form of under-citation, where relevant support is incomplete, or another support mismatch in which the necessary premises appear elsewhere in the original problem.

\paragraph{Example of under-citation.}

Story~107, conclusion index~1, illustrates this distinction with a complete two-premise instance:

\begin{quote}

\textbf{Premise 1.} Heinrich Schmidt was a German politician.
\[
\mathrm{German}(\mathrm{heinrichSchmidt})
\land
\mathrm{Politician}(\mathrm{heinrichSchmidt})
\]

\textbf{Premise 2.} Heinrich Schmidt was also a member of the Prussian State Parliament and the Nazi Reichstag.
\[
\begin{aligned}
&\mathrm{Member}(\mathrm{heinrichSchmidt},\mathrm{prussianStateParliament})\\
&\quad\land
\mathrm{Member}(\mathrm{heinrichSchmidt},\mathrm{naziReichstag})
\end{aligned}
\]

\textbf{Conclusion.} Some German politician was part of both the Prussian State Parliament and the Nazi Reichstag. \emph{(True)}
\[
\begin{aligned}
\exists x\,
\big(
&\mathrm{German}(x)
\land
\mathrm{Politician}(x)\\
&\land
\mathrm{Member}(x,\mathrm{prussianStateParliament})\\
&\land
\mathrm{Member}(x,\mathrm{naziReichstag})
\big).
\end{aligned}
\]

\end{quote}

The human-written derivation contains one step for this conclusion but names only premise~2 as support. Premise~2 establishes both membership relations but does not establish that Heinrich Schmidt is German or a politician; premise~1 establishes these properties but contains no membership information. Consequently,

\[
P_2 \not\models c_1,
\qquad
P_1 \not\models c_1,
\qquad
P_1 \land P_2 \models c_1,
\]

where $c_1$ denotes the FOL formula representing the step conclusion of the single derivation step. The translated claim is therefore entailed by the complete original premise set, but not by the cited premise set. This is a direct case of under-citation and illustrates why checking entailment against the stated support is stricter than checking whether the conclusion can be derived somewhere from the full problem context.

\subsection{Derivation-Level Checking Outcomes}
\label{sec:results-certification}

Applying the checking criteria defined in Section~\ref{sec:method} to the 386 complete derivations, we find that 202 (52.3\%) meet all criteria. Among the remaining 184 derivations, 173 contain at least one step that does not pass the checking procedure, while the other 11 fail to establish the required final target despite having no failed checked step.

The 202 derivations meeting all criteria form a reference subset that passes our entailment checks under the released FOL representations. None relies on a repaired FOL annotation, an automatically proposed symbol substitution, a manually corrected annotation, or a manually altered step translation. These results describe the logical support of the translated claims, rather than guaranteeing the correctness of the original natural-language derivations.

The proportion of derivations meeting our checking criteria should not be interpreted as a measure of human reasoning accuracy. Failure may arise from incomplete support, missing derivation references, structural problems, limitations of the available FOL vocabulary, or errors introduced during automatic translation. This distinction is particularly important given that 191 of the 230 translated claims not entailed by their cited premise sets are entailed by the complete original premise set. In addition, the procedure evaluates entailment rather than the inference-rule label supplied with a derivation step. A derivation passing all criteria should therefore be understood as one whose translated steps satisfy the specified entailment checks, while a derivation that does not pass is not necessarily logically incorrect.

\subsection{Manual Audit of Translation Faithfulness}
\label{sec:translation-audit}

To examine whether successful entailment checks rely on faithful formalizations, we manually inspected 50 randomly sampled steps that had passed the Z3 entailment checks. Of these, 46 translations (92\%) were classified as faithful, with an approximate 95\% Wilson confidence interval of 81.2--96.8\%. Three translations (6\%) were classified as stronger than their natural-language claims, while one (2\%) changed the intended meaning through incorrect symbol usage. No weaker translations were identified in the sample. The observed mismatches included interpreting an inclusive disjunction as exclusive, introducing an additional conjunct, changing a predicate into a constant, and broadening the scope of quantification. A genuinely stronger translation does not itself create a false pass for the original claim, provided that its cited support is faithfully represented. However, stronger translations may affect subsequent entailment checks if they are used as support for later steps. The remaining mismatch involved treating a predicate as an individual constant, potentially producing a misleading pass. Thus, passing an SMT-based entailment check does not guarantee translation faithfulness. The audit provides a preliminary assessment of translation reliability among passed steps rather than a guarantee of correctness across the full dataset.

\section{Analysis}
\label{sec:analysis}

\subsection{Interpreting Failed Entailment Checks}
\label{sec:analysis-failures}

The step-level results show that a failed entailment check should not be interpreted immediately as evidence that the corresponding human-written derivation step is logically incorrect. Of the 1{,}305 translated derivation steps evaluated by Z3, 230 are not entailed by their stated support. However, when these claims are checked again against the complete original premise set, 191 of the 230 become entailed. The remaining 39 are not entailed by the complete original premise sets.

This distinction separates failures in the recorded support from failures that persist at the level of the claim itself. In many cases, the translated claim is entailed by the complete original premise set, although its cited premise set does not entail it. Story~107 illustrates the clearest form of this problem: the claim requires information distributed across two original premises, while the human-written derivation cites only one of them. The failure therefore concerns the relationship between the derivation step and its stated support rather than the logical availability of the claim within the original problem.

This result is important because our checking criterion treats a human-written derivation as an account of why a conclusion follows, not merely as a sequence of claims that happen to be derivable somewhere from the complete problem context. A derivation step can therefore fail its cited-premise entailment check even when its translated claim is entailed by the complete original premise set. In this sense, step-level entailment checking evaluates both the logical content of the translated claim and whether the dependencies recorded in the derivation provide an adequate justification for that claim.

The 191 recovered cases show that most failed cited-premise checks concern translated claims that are nevertheless entailed by the complete original premise set. Instead, they reveal a mismatch between a derivation step and its stated support. The remaining 39 cases require a different analysis because their translated claims are not entailed by the complete original premise set. These cases cannot be explained solely by incomplete stated support and therefore require examination of the human-written derivation, the FOL translation, and the released FOL annotations.

\subsection{Manual Analysis of Claims Not Entailed by the Original Premises}
\label{sec:analysis-manual}

We manually examine the 39 derivation steps whose translated claims are not entailed by the complete original premise set under the released FOL annotations. For each case, we compare the natural-language derivation step, its automatically produced FOL translation, the relevant natural-language premises, and the corresponding released FOL annotations. This comparison identifies three main sources of failure: \emph{human reasoning errors}, \emph{FOL translation errors}, and \emph{dataset annotation errors}.

Table~\ref{tab:manual-errors} summarizes the resulting classification. Human reasoning errors account for 21 of the 39 cases, FOL translation errors for 11, and dataset annotation errors for 7.

\begin{table}[!htb]
\centering
\small
\begin{tabular}{lr}
\toprule
Source of failure & Steps \\
\midrule
Human reasoning error & 21 \\
FOL translation error & 11 \\
Dataset annotation error & 7 \\
\midrule
Total & 39 \\
\bottomrule
\end{tabular}
\caption{Manual classification of the 39 translated claims not entailed by the complete original premise set under the released FOL annotations.}
\label{tab:manual-errors}
\end{table}

\paragraph{Human reasoning error.}

A human reasoning error occurs when the claim made by the human-written derivation step does not follow from the available information, while the automatically produced FOL translation preserves the content of that claim. These failures include steps that omit a required condition, introduce information that is not available from their support, or rely on an earlier derivation step containing an invalid human inference.

Story~353, conclusion index~2, provides a direct example.

\begin{quote}

\textbf{Cited premise 6.} It is not true that ``As it was'' by Harry Styles is classical music that Max listens to and is from the 12th century.
\[
\begin{aligned}
\neg
\big(
&\mathrm{ClassicalMusic}(\mathrm{asItWasByHarryStyles})\\
&\land
\mathrm{Song}(\mathrm{asItWasByHarryStyles})\\
&\land
\mathrm{From}(\mathrm{asItWasByHarryStyles},\mathrm{12thCentury})
\big).
\end{aligned}
\]

\textbf{Human-written derivation step.}

{\ttfamily\raggedright D1: From (6): "As it was" by Harry Styles is neither classical music that Max listens to, nor is from the 12th century. [De Morgan's]\par}
\smallskip

\textbf{Automatically produced FOL translation.}
\[
\begin{aligned}
&\neg
\big(
\mathrm{MaxListenTo}(\mathrm{asItWasByHarryStyles})\\
&\qquad\land
\mathrm{ClassicalMusic}(\mathrm{asItWasByHarryStyles})
\big)\\
&\land
\neg
\mathrm{From}(\mathrm{asItWasByHarryStyles},\mathrm{12thCentury}).
\end{aligned}
\]

\end{quote}

The released FOL annotation for premise~6 does not fully preserve its natural-language wording: the phrase ``Max listens to'' is represented using the predicate \(\mathrm{Song}\), whereas the automatically produced translation of D1 uses \(\mathrm{MaxListenTo}\). This mismatch, however, is not what causes the verification failure in this case. When the claim is checked against the complete original premise set, the first part of D1, stating that the song is not classical music that Max listens to, is supported by the available premises. The component not entailed by the original premises is the additional claim that the song is not from the 12th century.

Premise~6 states only that the conjunction of the relevant properties does not hold. From
\[
\neg(P \land Q \land R)
\]
it follows that
\[
\neg P \lor \neg Q \lor \neg R,
\]
but not that any particular member of the conjunction is individually false. D1 therefore strengthens the premise by asserting that the song is not from the 12th century, which is not licensed by the available information. The failed entailment consequently results from the human-written reasoning rather than from the vocabulary mismatch in the formal representation.

\paragraph{FOL translation error.}

A FOL translation error occurs when the natural-language derivation step has a defensible interpretation under its stated support, but the automatically produced FOL translation does not preserve that interpretation. These cases include changes in quantification, logical scope, predicate use, or the interpretation of logical connectives.

Story~200, conclusion index~0, illustrates this distinction.

\begin{quote}

\textbf{Cited premise 1.} Wildfeed exists as an unannounced program.
\[
\exists x\,
\big(
\mathrm{Wildfeed}(x)
\land
\mathrm{Unannounced}(x)
\land
\mathrm{Program}(x)
\big).
\]

\textbf{Human-written derivation step.}

{\ttfamily\raggedright D1: From (1): Wildfeed is a program. [conjunction elimination]\par}
\smallskip

\textbf{Automatically produced FOL translation.}
\[
\mathrm{Program}(\mathrm{Wildfeed}).
\]

\textbf{Faithful FOL representation.}
\[
\exists x\,
\big(
\mathrm{Wildfeed}(x)
\land
\mathrm{Program}(x)
\big).
\]

\end{quote}

In the released annotation, \(\mathrm{Wildfeed}\) is used as a predicate rather than as an individual constant. The automatic translation instead treats \(\mathrm{Wildfeed}\) as a constant and therefore produces a formula that is not entailed by premise~1. When the step is represented using the same predicate structure as the original annotation, the claim follows directly from the cited premise. The failed entailment check therefore results from the FOL translation rather than from the human-written reasoning.

Across the 11 cases classified as FOL translation errors, manual correction of the formal representation is sufficient for the corresponding claim to pass the entailment check. These corrections are used only for the manual diagnosis of the failure source and are not applied to the main working set or to the reported checking outcomes.

\paragraph{Dataset annotation error.}

A dataset annotation error occurs when the released FOL annotation does not preserve information expressed in the corresponding natural-language premise. In this case, both the human-written derivation step and its automatically produced translation can be consistent with the natural-language problem, while the entailment check fails because the released formal representation has omitted or changed part of that information.

Story~55, conclusion index~1, provides an example.

\begin{quote}

\textbf{Cited premise 6.} Bob is a businessman who has ownership over Microsoft.
\[
\mathrm{BusinessPerson}(\mathrm{bob})
\land
\mathrm{Ownership}(\mathrm{bob},\mathrm{microsoft}).
\]

\textbf{Human-written derivation step.}

{\ttfamily\raggedright D1: From (6): Bob is a businessman. [conjunction elimination]\par}
\smallskip

\textbf{Automatically produced FOL translation.}
\[
\mathrm{BusinessMan}(\mathrm{bob}).
\]

\end{quote}

The predicate \(\mathrm{BusinessMan}\) is already part of the problem vocabulary: premise~3 states that every business person is either a businessman or a businesswoman. The natural-language premise explicitly states that Bob is a businessman, so the human-written derivation step is supported by the natural-language text. However, the released FOL annotation represents Bob as a \(\mathrm{BusinessPerson}\) and does not encode the predicate \(\mathrm{BusinessMan}(\mathrm{bob})\). Consequently, Z3 cannot establish the translated derivation step from the released annotation. The failure therefore originates in the mismatch between the natural-language premise and its dataset-provided FOL annotation.

These examples illustrate why the 39 remaining failures cannot be treated as a single class of incorrect derivation steps. A failure may arise from the reasoning expressed by the human-written derivation, from the translation of that reasoning into FOL, or from the released formal representation of the original problem. More specific phenomena, such as omitted conditions, incorrect quantifier scope, predicate--constant confusion, or inherited reasoning errors, can be treated as subcategories within these three sources of failure.

\subsection{Implications for Reference Reliability and Epistemic Risk}
\label{sec:analysis-implications}

The results have a direct implication for the use of human-written derivations as references in reasoning evaluation. Human authorship alone does not guarantee that a derivation provides a complete and formally supported account of why a conclusion follows. Within the working set, verification failures arise at several points in the reference: a derivation step may cite insufficient support, the human-written claim itself may not follow, the automatic FOL translation may alter the meaning of the claim, or the released FOL annotation may not preserve the corresponding natural-language problem.

The distinction between these sources is particularly important when a human-written derivation is used as a reference against which another reasoning process is evaluated. A disagreement with such a reference is not automatically evidence of an error in the system being evaluated. If the human-written derivation contains an invalid inference, if its cited premise set is incomplete, or if the underlying FOL annotation is mismatched with the natural-language problem, the observed disagreement may originate partly or entirely in the reference. Assessing reference reliability therefore becomes part of the evaluation process itself rather than a separate assumption made before evaluation begins.

The 191 claims recovered by the complete original premise set provide a particularly clear example of this problem. Their translated claims are entailed by the complete original premise sets, but not by their cited premise sets. Treating all 230 initial failures as human reasoning errors would therefore substantially mischaracterize the data. Conversely, the manual analysis of the remaining 39 claims shows that even a failure under the complete original premise set is not sufficient to identify its source automatically. Human reasoning, automatic formalization, and dataset annotation must be distinguished before a failure can be interpreted.

The derivation-level results should be understood in the same way. The 202 derivations meeting all our checking criteria have translated steps that pass the specified entailment checks and establish the required target under the available FOL representations. The remaining 184 derivations do not meet all criteria, but this does not imply that all of them contain incorrect human reasoning. Their failures include incomplete stated support, missing derivation references, structural problems, unavailable FOL representations, translation errors, and annotation mismatches. The 52.3\% pass rate therefore indicates how much of the working set satisfies the complete checking procedure, rather than measuring human reasoning accuracy.

These findings illustrate the \emph{epistemic risk} defined in the Introduction. Errors in human-written derivations or their FOL annotations may become accepted as unquestioned ground truth, potentially undermining the reliability of knowledge claims derived from the dataset. Our analysis identifies possible sources of this risk within the reference representations themselves. However, we do not directly measure its consequences for downstream AI evaluation.

Although the present analysis is instantiated on P-FOLIO, this implication is not specific to that dataset. Any evaluation framework that relies on intermediate reference reasoning inherits a similar dependency on the quality of the reference representations and the relationships between them. Where the necessary components are available, step-level SMT-based entailment checking provides one way to examine the logical support of the reference separately from the behaviour of the system evaluated against it.

\subsection{Limitations and Future Work}

The present study has several limitations. First, although our method is not conceptually restricted to P-FOLIO, its implementation relies on FOLIO-style notation and is evaluated only on instances labelled True or False. Second, LLM-generated FOL translations may not faithfully preserve the original natural-language claims. Our manual audit of 50 passed steps provides preliminary evidence about translation faithfulness, but does not establish correctness across the full dataset. Third, the entailment checks depend on the released FOL annotations, which may themselves contain errors. We exclude cases requiring annotation repair and do not independently verify every formal representation. Finally, our method checks logical entailment rather than the correctness of the inference-rule labels assigned by human annotators.

Importantly, successful entailment checks do not constitute end-to-end verification of natural-language reasoning. We do not quantify epistemic risk or directly measure how the identified reference problems affect downstream AI evaluation. Our findings should therefore be interpreted as an empirical assessment of logical support under the available FOL representations.

Future work could extend the method to other reasoning datasets, improve translation reliability through multiple formalizers or independent manual validation, and examine inference-rule labels alongside logical entailment. Applying the approach to model-generated reasoning could also help investigate how reference-level errors influence reasoning evaluation.

\section{Conclusion}

This study examined the logical support of human-written reasoning chains used as references for evaluating generative AI reasoning. Using P-FOLIO, we developed an entailment-checking pipeline that translates natural-language derivation steps into first-order logic and checks whether each translated claim follows from the premises or earlier derivations explicitly cited by the annotator. The results show that establishing the problem-level target does not guarantee that the FOL formula for each step conclusion is entailed by its cited premise set. Many failed step-level checks are explained by incomplete or mismatched support, while the remaining failures arise from distinct sources including human reasoning errors, automatic FOL translation errors, and mismatches in the released dataset annotations.

These findings suggest that human-written derivations should not automatically be treated as reliable reference explanations. Errors in their step conclusions, cited premise sets, or FOL representations may remain unnoticed when the dataset is treated as ground truth. Our entailment-checking method helps identify such mismatches and distinguish their possible sources. This addresses an important aspect of epistemic risk in reasoning datasets, while the effects of these errors on downstream AI evaluation remain for future research.

\bibliographystyle{abbrvnat}
\bibliography{bibliography_ai_safety}
\clearpage

\appendix

\section{Implementation Details}
\label{app:implementation}

\paragraph{Translation.}
The FOL translation stage used \texttt{openai/o3-mini} through the OpenRouter API.
The cached translations used in the reported experiments were generated on or
before 14 July 2026. Each derivation was translated as a single request
containing all of its derivation steps. We used a temperature of 0 and a
maximum output length of 4000 tokens. Reasoning effort was not explicitly
configured and therefore followed the provider default.

Requests were not resampled for alternative translations; retries occurred only
after an API error or a truncated response, with up to six retries. The first
successful response was cached and used in the subsequent experiments.

\paragraph{Preprocessing.}
Before parsing, Unicode and ASCII logical operators were normalized, accents
were removed, and trailing periods were stripped. Identifiers beginning with a
digit were prefixed with \texttt{n}, and a missing quantified variable was
replaced with \texttt{x}. All formulas were represented using a single
uninterpreted sort, and constants were kept distinct. A translation that could
not be parsed after preprocessing was recorded as having no usable FOL
translation.

\paragraph{Entailment checking.}
Entailment checking used \texttt{z3-solver} version 4.16.0.0 with a timeout of
5000\,ms per satisfiability query. The stated support was first checked for
satisfiability on its own; unsatisfiable support was marked as vacuous. For
satisfiable support, entailment of the step-conclusion formula \(c_i\) from
its cited premise set \(S_i\) was tested by checking whether
\(S_i \cup \{\neg c_i\}\) was unsatisfiable. An \texttt{unsat} result was treated as entailed,
\texttt{sat} as not entailed, and \texttt{unknown} as undecided. An undecided
result prevents a derivation from meeting the checking criteria. No Z3 query, including the problem-level label checks and the step-level
entailment checks, returned \texttt{unknown}.

\section{Translation Prompt}
\label{app:translation-prompt}

The following prompt was used for the automatic translation of human-written
derivation steps into FOL. The placeholders were replaced with the premises,
conclusion, and complete derivation of each problem instance.

{\small
\begin{verbatim}
You are a faithful logical transcriber. Below are the first order logic
premises of a reasoning problem and the derivation steps. Translate each
step's derivation sentence into exactly one first order logic formula.

Transcribe EXACTLY what each step says. Do NOT solve the problem, do NOT
correct steps that look logically wrong, do NOT complete missing reasoning.
A wrong step must become the formula for the wrong statement it makes.

Vocabulary rules:
- Use ONLY the predicate names and constant names that appear in the premises
  (and conclusion) below, with identical spelling and arity.
- Do not invent new predicates or constants. If a step talks about something
  this vocabulary cannot express, output an empty string "" for that step's
  formula.
- Use this notation exactly: FORALL and EXISTS spelled with the symbols shown
  in the premises, and the connectives shown there (implication arrow,
  conjunction, disjunction, negation, exclusive or, biconditional, equality).
  Write quantified variables as x, y, z.

Formula rules:
- One self-contained formula per step, expressing the step's stated conclusion
  (the fact it claims), not its justification.
- If a step states something about a specific named entity, write a ground
  formula about that constant, not a quantified one.
- If a step restates a premise, its formula will simply match that premise.
  That is correct transcription.

Return ONLY a JSON object of this exact shape:
{
  "steps": [
    {"step_id": "D1", "fol": "..."},
    {"step_id": "D2", "fol": "..."}
  ]
}

PREMISES (first order logic):
{premises_fol}

CONCLUSION (first order logic, for vocabulary only):
{conclusion_fol}

DERIVATION STEPS TO TRANSCRIBE:
{steps_block}
\end{verbatim}
}

\end{document}